\documentclass{article}
\usepackage{arxiv}
\usepackage{natbib} %
\usepackage{caption} %
\usepackage{hyperref}

\usepackage[textsize=tiny]{todonotes}
\usepackage{amsmath}
\usepackage{amssymb}
\usepackage{mathtools}
\usepackage{amsthm}

\usepackage[capitalize,noabbrev]{cleveref}

\theoremstyle{plain}
\newtheorem{theorem}{Theorem}[section]

\theoremstyle{definition}
\newtheorem{definition}[theorem]{Definition}

\theoremstyle{remark}

\usepackage[textsize=tiny]{todonotes}

\usepackage{amsmath, amssymb}
\usepackage{xcolor}
\usepackage{graphicx}
\usepackage{soul}
\usepackage{algorithm}
\usepackage{multirow}
\usepackage{array}
\usepackage{xcolor, colortbl}
\usepackage{subfigure}

\newcommand{\mc}{memory classifier}
\newcommand{\mcs}{memory classifiers}
\newcommand{\reals} {\mathbb{R}}

\newcommand{\score} {\mathbf{sim}}
\newcommand{\s} {\mathbf{sim}}

\definecolor{sd}{RGB}{0,0,255}

\definecolor{ed}{RGB}{225,0,0}

\newcommand{\ep}{\varepsilon}

\def\eg{\textit{e.g., }}
\def\ie{\textit{i.e., }}
\def\etc{\textit{etc. }}

\definecolor{yy}{RGB}{0,120,120}

\usepackage{algorithmic} %

\title{Memory Classifiers: Two-stage Classification for Robustness in Machine Learning}
\renewcommand{\shorttitle}{Memory Classifiers: Two-stage Classification for Robustness in Machine Learning}

\date{}
\renewcommand{\undertitle}{}
\author{%
  Souradeep Dutta\\
  PRECISE Center\\
  Computer and Information Science\\
  University of Pennsylvania\\
  \texttt{duttaso@seas.upenn.edu}
  \And 
  Yahan Yang \\ 
  PRECISE Center\\
  Electrical and Systems Engineering\\
  University of Pennsylvania\\
  \texttt{yangy96@seas.upenn.edu}
  \And 
  Elena Bernardis \\ 
  Department of Dermatology\\
  Perelman School of Medicine\\
  University of Pennsylvania\\
  \texttt{elber@seas.upenn.edu}
  \And 
  Edgar Dobriban\\
  Statistics and Data Science\\
  Wharton School of Business\\
  University of Pennsylvania\\
  \texttt{dobriban@wharton.upenn.edu} \\
   \And 
  Insup Lee\\
  PRECISE Center\\
  Computer and Information Science\\
  University of Pennsylvania\\
  \texttt{lee@seas.upenn.edu} \\
}

\begin{document}
\maketitle

\begin{abstract}
The performance of machine learning models can significantly degrade under distribution shifts of the data. We propose a new method for classification which can improve robustness to distribution shifts,
by combining expert knowledge about the ``high-level" structure of the data with standard classifiers. 
Specifically, we introduce two-stage classifiers called \textit{memory classifiers}. First, these identify prototypical data points---\textit{memories}---to cluster the training data. This step is based on features designed with expert guidance; for instance, for image data they can be extracted using digital image processing algorithms. Then, within each cluster, we learn local classifiers based on finer discriminating features, via standard models like deep neural networks. We establish generalization bounds for memory classifiers.
We illustrate in experiments that they can improve generalization and robustness to distribution shifts on image datasets. We show improvements which push beyond standard data augmentation techniques. 
\end{abstract}

\section{Introduction}
\label{sec:introduction}

Modern machine learning (ML) systems have achieved extraordinary performance in several areas, such as computer vision and natural language processing. However, recent work has illustrated that seemingly benign data corruptions, that humans can readily withstand, are potentially detrimental to the efficacy of ML systems \citep{cifar10_fails, transforms_fail}. For vision, this has been observed for corruptions like pixelation, compression, blur, and other transforms \citep{hendrycks}. This lack of robustness has limited the application of ML techniques in safety-critical applications. 
Efforts to improve the robustness against such corruptions have been attempted in the form of data augmentation \citep{hendrycks2019augmix, noisy_mix}. However, many systems remain vulnerable even to small degrees of corruption.
An alternative classical approach is to infuse ML models with appropriate inductive biases \citep{prior_knowledge_2, prior_knowledge_4, prior_knowledge_3}. Such priors are available in at least two distinct ways: as model architectures such as deep neural nets (DNNs) \citep{fukushima1980neocognitron,lecun1989backpropagation} and as synthetic examples \citep{ai_system_critic}. 

\begin{figure}[t!]
    \centering
    \includegraphics[width = 13.5cm, height = 3.5cm]{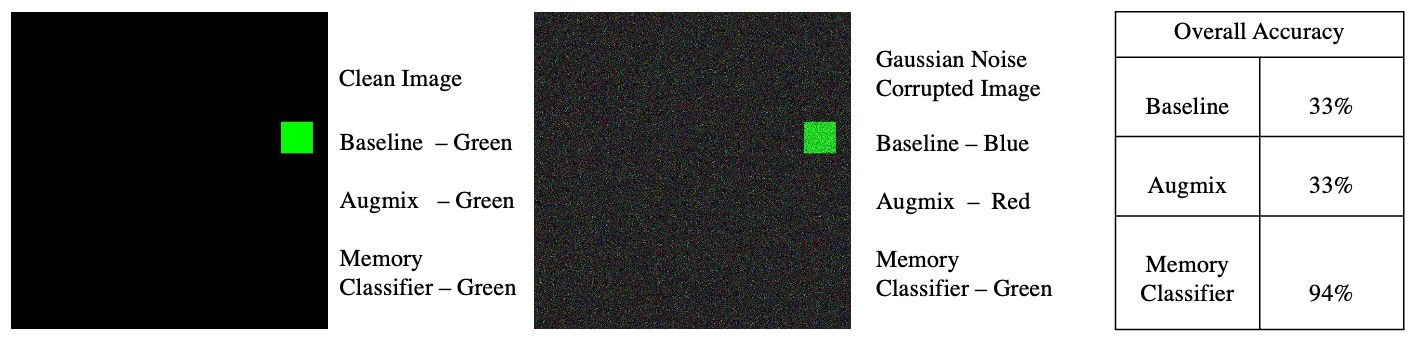}
    \caption{\footnotesize The figure shows sample images with the corresponding labels assigned by different classifiers. The image in the middle equals the one on the left plus Gaussian noise. We additionally show the classifier accuracies under Gaussian noise with severity level $5$ \citep{hendrycks} }
    \label{fig:motivating_example}
\end{figure}

As a motivating example, consider $L \times L$ images generated as follows. For $w \in (0, L)$, consider a random vector $(X,Y)\sim P$ with a probability density function given for $(x,y) \in  [w/2 , L-w/2 ]^2$  by
$p(x,y) = 
(L-w)^{-2}$, and zero otherwise.
Consider $\alpha \in \reals^3$, and generate an image vector $I_{(\alpha,x,y)}\in\reals^{L \times L \times 3}$ as
\begin{equation*}
\begin{aligned}
I_{(\alpha,x,y)} [i,j, :] = 
\begin{cases}
\alpha,&  i \in (x - w/2 , x + w/2 ) \text{ , } j \in (y - w/2 , y + w/2 ), \\
(0, 0, 0),& \text{otherwise}.
\end{cases}
\end{aligned}
\end{equation*}

Hence, for an $8$-bit RGB color scheme, $\alpha = (255, 0, 0)$ generates images with a red patch of size $w \times w$, inside an $L \times L$ image whose pixels are sampled from the distribution $P$. 
Now consider a toy color dataset $\mathcal{D}$ with three distinct classes corresponding to the color channels generated by the above process. 
The dataset contains $3\times 10^3$ train and $3\times 10^2$ test datapoints. 
The task is to predict the color of the patch contained in the image.
We train a SOTA CNN architecture ResNet50, and---as expected given the simplicity of the dataset---it achieves $100\%$ test accuracy.  However, perhaps surprisingly, these classifiers are not robust to corruptions like Gaussian noise \citep{hendrycks}, even on this simple dataset; see Figure \ref{fig:motivating_example}. 

A standard way to improve accuracy against such corruptions is through data-augmentation methods such as AugMix \citep{hendrycks2019augmix}. We train a ResNet50 architecture on the color dataset using AugMix. The prediction accuracy on the perturbed test dataset improves, but remains well below satisfactory levels. 
For instance, under Gaussian noise, the accuracy of the network trained with AugMix drops to $33\%$, 
which is close to random guessing, as this is a balanced three-class dataset. The details of this experiment are in the Appendix \ref{sec:color-dataset-results}. 

This task does not pose problems to humans,
as the color dataset has a clear feature---color of the patch---that one can use.
However, deep nets fail spectacularly, and thus seemingly do not learn this obvious feature. 
Our experiments show that this is the case in more realistic image datasets for high stakes medical applications. Motivated by these examples,  in this paper we aim to address the following research question: \\

\textit{How do we introduce expert information into classifiers, to improve robustness without compromising performance?} \\

To this end we introduce \textit{memory classifiers}. In this framework, we first allow the designer to write a distance function which captures a notion of similarity among images. This step is particularly useful in applications such as medical diagnosis, where physicians often have useful intuitions to share, and we illustrate this in one of our case studies. Next, using the distance metric we cluster the dataset around training datapoints and treat them as \textit{memories}. Afterwards, we assign a cluster to each test datapoint according to the most ``similar" memory, and apply classifier trained for the cluster. The distance function in the case for image applications, we leverage the rich body of work on digital image processing, \eg \cite{gonzalez2009digital}. This offers us tools to capture the high-level properties of an image, which can be potentially more robust.

We provide both theoretical and empirical evidence to support 
the benefits of this architecture. 
We develop generalization bounds, and  
empirically show that it has robustness to small distribution shifts. Intuitively, the memory-formation step identifies representative datapoints that are well-separated in the space of datapoints, which reduces sample complexity and increases robustness.


\textbf{Contributions:}
1. We propose \textit{memory classifiers}, a new two-stage method for classification.
The first step is to select representative data points, or \emph{memories}, from the training dataset,
representing different clusters of the data.  We learn a classifier for each such cluster. To select memories, we propose a method to encode domain knowledge via a similarity metric.   
2. We give rigorous generalization bounds for memory classifiers.
3. We show improvements in classification robustness across different types of ML architectures, by evaluating robustness against natural distribution shifts on two image datasets. Further, any technique that improves the robustness of a neural network architecture can be incorporated  into the within-cluster classifiers.


\section{Overview}
\begin{figure*}[!th]
    \centering
    \includegraphics[height = 7cm, width = 15cm]{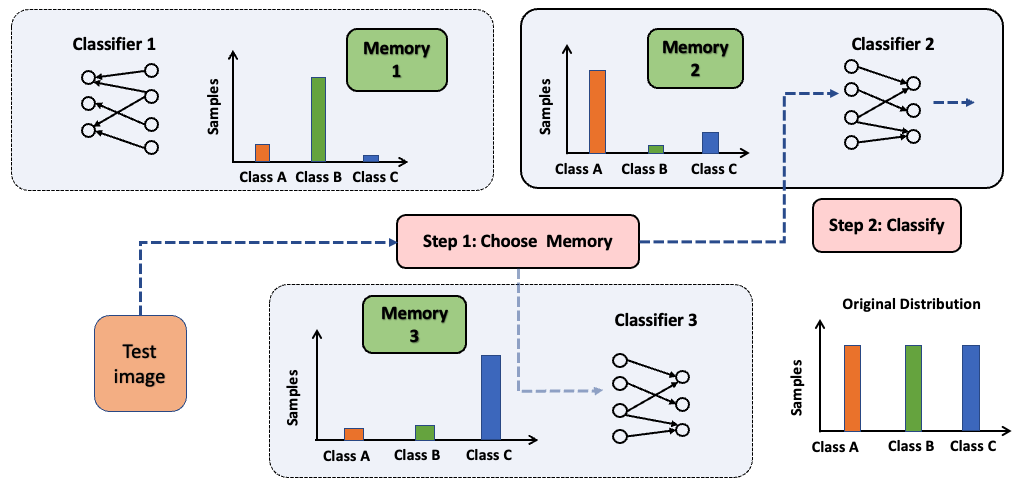}
    
    \caption{\footnotesize{\textit{Memory classifiers overview.} At test time, memory classifiers have \textbf{two} stages. Step 1: The test datapoint gets mapped to a \emph{memory} datapoint using a \emph{matching algorithm}. Step 2: Once matched, the final decision is issued by a \emph{within-cluster} classifier. 
    }}
    \label{fig:memory_classifier}
\end{figure*}
\label{sec:memory_classifier_overview}
Figure \ref{fig:memory_classifier} provides an overview for the computational workflow of memory classifiers. A memory classifier is composed of a collection of \textit{memories}, each equipped with a separate classifier. These individual classifiers could be of distinct types (DNNs, random forests, etc). Our two-stage memory classifier has the following steps:


\textbf{Step 1: Memory Selection.} 
The memory selection algorithm maps a datapoint to a memory $m$ using a similarity score $\score$.  A higher value of $\score(x_1,x_2)\in  [0,1]$  means that the datapoints $x_1$ and $x_2$ are more similar. Formally, given a set of memories $M^+= \{ (m_1, b_1), ( m_2, b_2), \dots, (m_q, b_q) \}$, where $b_k\in[0,1]$ are similarity score thresholds, and a test datapoint $x_t$, we find the closest memory which satisfies the threshold condition by computing: 
\begin{equation}
    \label{eq:nearest_mem}
    ( m_k, b_k) \in \underset{( m_i, b_i) \in M^+}{\arg\max \text{ } } \score(m_i, x_t).
\end{equation}
If $\score( m_k, x_t) \ge b_k$, we then select the classifier associated with memory $m_k$, otherwise we select an ``out-of-boundary" classifier.\footnote{If needed, ties can be handled arbitrarily, for instance by assigning $x_t$ to the memory $m_i$ with the smallest index $i$ that achieves the maximum.} 

\textbf{Step 2: Classification.} 
Use $x_t$ as an input to the classifier associated with memory $m_k$, to predict a label. If the ``out-of-boundary" classifier is selected,  label the input $x_t$ as an unknown class. 


\subsection{Memory Classifiers: Formal Analysis}

Consider a selector function $s : \mathcal{X} \times [q+1]:=\{1,2,\ldots,q+1\} \rightarrow \{0,1\}$, which chooses the $k$-th memory for input $x$ when $s(x,k)=1$. 
Further, $s$ chooses the out-of-distribution classifier when $s(x,q+1)=1$. 
A memory classifier is then defined by the function $F:\mathcal{X}\to \mathbb{R}$, which first uses $s$ 
to select a memory $k$
and then applies a classifier $h_k$, so that:
\begin{equation}
    \label{eq:cascade_function}
        F(x) = \sum_{k = 1}^{q+1}s(x,k)h_k(x).
\end{equation}
Here 
$h_k$ is the within-cluster classifier associated with the chosen memory $k$, for $k\le q$, and $h_{q+1}$ is the out-of-boundary classifier. 
We classify $x$ according to the sign of $F(x)$, breaking ties arbitrarily.

Now, we provide a data-dependent generalization bound for \mcs. 
We let the training and testing samples be independently and identically distributed (iid), sampled
from a fixed unknown distribution $D$ over the space $\mathcal{X} \times\{-1, +1\}$. 
For a binary label $y \in \{-1, 1\}$, we denote the classification error of a classifier $F$ as
$R(F) = P_{(x,y) \sim D}[yF(x) \leq 0] $. The empirical error over a training set 
$T = \{(x_1, y_1), \dots, (x_n, y_n)\}$ is  $\smash{\hat{R}(F) = P_{T}[yF(x) \leq 0}]$, where $P_T$ denotes the empirical distribution defined by $T$. 
We assume that for all $k\in[q+1]$, $h_k$ are chosen from a hypothesis space $H_k$, while $s$ is chosen from a hypothesis space $\mathcal{S}$. 
Let the number of points mapped to memory $k$ and correctly classified be $n_k^+$. 
Let the empirical Rademacher complexities 
\citep{koltchinskii2002empirical,bartlett2002rademacher,shalev-shwartz_ben-david_2014}
of the hypothesis spaces $H_k,\mathcal{S}$ over the training set $T$ be $\smash{\hat{\mathfrak{R}}_T(H_k), \hat{\mathfrak{R}}_T(\mathcal{S})}$, respectively.
Consider  $0<\delta \le 1/q$, and let
\begin{equation}\label{C}
    C(n,q,\delta)= 
    \frac{1}{\delta} \sqrt{\frac{\log q}{n}}
    \left(2 + \sqrt{\log\frac{\rho^2 n}{\log q}}\right)
    + \sqrt{\frac{\log(4/\delta)}{n}}.
\end{equation}
Assume that the memories are chosen from the training datapoints, so $M\subset \{x_1,\ldots,x_n\}$. This always holds in our experiments. 
Then, for any memory classifier, with probability at least $1-\delta$, we have for some constant $\kappa>0$,
\begin{equation}
\begin{aligned}
    R(F) & \leq \hat{R}_T(F) + 
     4q\left[\frac{q (1 + \log n)+\kappa}{n^{1/2}}+\max_{k=1}^{q} \hat{\mathfrak{R}}_T(H_k) 
     \right] + C(n,q,\delta).
\end{aligned}
\label{eq:generalization_bound}
\end{equation}
The details of the proof are in Appendix \ref{sec:memory_classifier_theory}, and build on the results of \citep{deep_cascade}. The proof also requires a delicate bounding of the number of possible selector functions induced by memory classifiers; which requires a careful combinatorial analysis, and also uses the Massart Lemma from learning theory \citep{shalev-shwartz_ben-david_2014}.

In \eqref{eq:generalization_bound}, we have a nontrivial generalization bounds if the number $q$ of memories is of smaller order than $n^{1/4}/\log(n)^{1/2}$, and if the Rademacher complexity of each $H_k$ is $\hat{\mathfrak{R}}_T(H_k) \ll 1/ q$. 
Consistent with this, the memory selection algorithm \ref{alg:learn_memories} attempts to minimize the number of memories to achieve good generalization.
\section{Learning Memory Classifiers}

\label{sec:memory_classifier_learning}

Learning a memory classifier has two stages. 
In the first stage, we identify datapoints in the training data as memories, such that all data points are within a certain similarity threshold 
of the memories, while aiming to minimize the number of memories.
In the second stage, we train a classifier for data cluster belonging to each memory. We explain the two steps in detail next.\\
 \textbf{Clustering with Memories:} 
We aim to cluster the data around the memories, which should be datapoints in the training set. 
Given a dataset, the \emph{Partitioning Around Medoids} (PAM) method \citep{park2009simple,xu2015comprehensive} tries to select a cluster $ M=\{m_1, m_2, \dots , m_q \}$ of $q$ memories such that the following clustering objective is maximized: $\sum_{i=1}^{n} \underset{m_j \in M}{\max}\s(m_j, x_i)$,
for the similarity score $\s:\mathcal{X}\times \mathcal{X} \rightarrow [0,1]$. 
The naive implementation of PAM has a runtime complexity of $O(n^2q^2)$ \citep{pam_compare}. 
Even the faster variants are usually too slow for large image datasets.
Thus, we use a simple variant of the  \emph{Clustering Large Applications based upon Randomized Search} (CLARANS) \citep{clarans} algorithm. This integrates randomized global search with local cluster improvement. 
\subsection{Selecting memories}

\label{sec:init_memories}

\textbf{Memory Initialization:} Here we provide a simple intuitive explanation of the memory set initialization workflow. The goal is to place memories in the input space densely enough, with the constraint that every training datapoint occurs within a similarity score boundary $b_t$ of some memory. The exact details are in Algorithm \ref{alg:learn_init_memories} in the Appendix \ref{sec:init_memories}. Essentially, we iterate the following process - a random datapoint is picked as a memory, and compared with all datapoints in the set in a single linear pass. The datapoints which end up being marked as similar are admitted onto the set of datapoints accepted by the current memory. 
This continues until the full set of training datapoints is covered by a set of memories $M$. This provides a warm start for any subsequent algorithm optimizing the choice of memories.

\label{sec:learning_memories}
\textbf{Learning Memories:} Starting with a dataset $\mathcal{D}$ of size $n$, the aim is to build a set $M = \{m_1, m_2, \dots, m_q\}$ of size $q$ which minimizes the clustering objective. Choosing the memories can be simplified by viewing it as a search through a graph $\mathcal{G}$ \citep{clarans} with subsets $\mathcal{D}_q \subset \mathcal{D}$ as its nodes. A subset of size $q$ defines a choice for the memory set $M$. 
\begin{definition}[\textbf{Memory Search Graph $\mathcal{G}$}]
The undirected memory search graph $\mathcal{G}$ is represented by an ordered pair $(V, E)$, where the set of nodes $V$ is the collection of subsets $\mathcal{D}_q$ of size $q$ of the original dataset $\mathcal{D}$. 
An edge $e \in E$ exists between two nodes $\mathcal{D}^1_q$ and $\mathcal{D}^2_q$ iff $|\mathcal{D}^1_q \cap \mathcal{D}^2_q| = q-1$. 
\end{definition}
Associated with each node of the graph is a clustering objective cost. This allows for a greedy algorithm: starting from a node one can visit a series of  neighboring nodes with increasing scores in the search process. This is  followed by a combination of $Global$ resets and $Local$ minimization to approximate the optimal choice.
\begin{algorithm}[h]
\caption{Generating Memories}
\label{alg:learn_memories}
\flushleft
\textbf{Input: } Dataset $\mathcal{D}= \{x_1, x_2, \dots, x_n \} $  \\
\textbf{Output: } Memories $M = \{ m_1, m_2, m_3, \dots, m_q \}  $ \\
\textbf{Parameters:} Max Global Steps $Z_g$, Max Local Steps $Z_l$, Similarity Score Threshold $b_t$
\begin{algorithmic}[1]
\STATE BestScore = $0$
\FOR{$ 1 \leq g \leq Z_g$}
\STATE Memory Set $M$ = GenerateInitialMemories($\mathcal{D}, b_t$)
\STATE $v$ = FindNode($M$, $\mathcal{G}$)
\STATE $\mathcal{G}$ = CreateGraph($\mathcal{D}$, $|M|$) 
\COMMENT{The memory search graph}
\STATE CurrentScore = ComputeScore($v$)
\FOR{$1 \leq l \leq Z_l $}
\STATE $v'$ = PickNeighbor($v, \mathcal{G}$)
\STATE NewScore = ComputeScore($v'$)
\IF{NewScore $>$ CurrentScore }
\STATE $v \leftarrow v'$
\STATE CurrentScore $\leftarrow$ NewScore
\ENDIF
\ENDFOR
\IF{CurrentScore $>$ BestScore}
\STATE BestScore = CurrentScore
\ENDIF
\ENDFOR
\STATE \OUTPUT $M$
\end{algorithmic}
\end{algorithm}

Algorithm \ref{alg:learn_memories} creates the memories which are used to implement the \mc.
Analogous to the standard \emph{CLARANS} algorithm \citep{clarans}, a node in $\mathcal{G}$ is surrounded by $q(n-q)$ neighbors, which can be large set
We warm start the process with a reasonable choice of an initial node in $\mathcal{G}$, and greedily search for local improvements for a fixed number of iterations. The number of memories is controlled by the similarity score thresholds $b_t$. The iterations are limited by $Z_g$ and $Z_l$.


 \subsection{Training a Classifier}
 For each memory, the training datapoints in the cluster associated to it form the training set for the classifier assigned to that memory. We can then use standard learning procedures to train these classifiers, as we explain in detail for each particular example.

\section{Experiments}
\label{sec:experiments}
We apply our framework to two classification tasks: acne lesion classification for dermatology and leaf severity assessment for botany. When applying machine learning tools to specific problems, there are often additional visual cues that can inform the decision of a human observer
and should ideally be useful to a machine learning algorithm. 
These features, if carefully extracted, can help improve the accuracy of classification; and possibly help in improving robustness as well---in line with the long history of feature engineering techniques \citep{feature_engineering_book}. 
The challenge is to incorporate them into a modern machine learning pipeline. 
\subsection{Case Study: Acne Lesion Classification}
\begin{figure}[h]
    \centering
    \includegraphics[height = 4.0cm, width = 14cm]{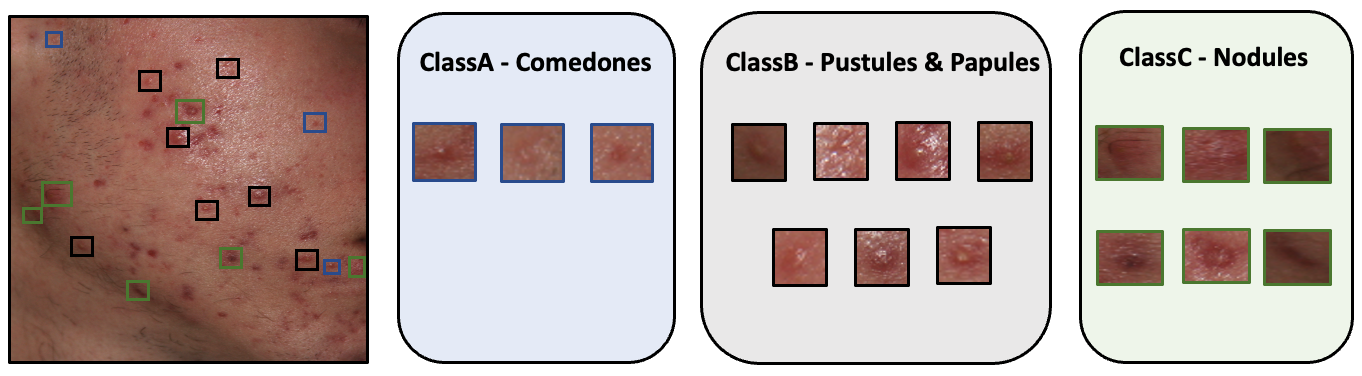}
\caption{\footnotesize \textit{Case Study: Acne lesion classification.} From each face image, we crop out primary acne lesions for classification. The lesion crop serves as an input to the classifier. The goal of the classifier is to predict the label of a lesion.}
\label{fig:acne_sample}
\end{figure}
\begin{table*}[ht]
\tiny
\begin{center}
\begin{tabular}{ p{1.5cm} p{0.5cm} p{1cm} p{0.5cm} p{1cm}  p{0.5cm} p{1cm} p{0.5cm} p{1cm} p{0.5cm} p{1cm}  }
    \hline
    & \multicolumn{2}{ c }{Adversarial} &  \multicolumn{2}{c}{AugMix} & \multicolumn{2}{c}{CutMix} &  \multicolumn{2}{c}{NoisyMix} & \multicolumn{2}{c}{ManifoldMix} \\
    
    \hline 
    & Baseline & Memory Classifier &  Baseline & Memory Classifier & Baseline & Memory Classifier &  Baseline & Memory Classifier &  Baseline & Memory Classifier\\
    \hline 
    Clean Accuracy & \textbf{65} & 63 & \textbf{68} & 65 & 49 & \textbf{66} & 48 & \textbf{63} & \textbf{76} & 71 \\
    \hline
    \hline
        Corruption &  \multicolumn{8}{c}{Average Accuracy} \\
    \hline
Brightness  & \textbf{60} & 56 & 63 & \textbf{69} & 53 & \textbf{64} & 45 & \textbf{59} & 53 & \textbf{58}\\ 
Contrast  & 63 & \textbf{67} & 63 & \textbf{76} & 58 & \textbf{73} & 57 & \textbf{72} & 64 & \textbf{70}\\ 
Elastic  & 56 & \textbf{60} & 55 & \textbf{66} & 52 & \textbf{63} & 48 & \textbf{62} & \textbf{69} & 67\\ 
Pixelate  & 48 & \textbf{55} & 60 & \textbf{68} & 52 & \textbf{65} & 48 & \textbf{63} & \textbf{74} & 70\\ 
JPEG  & 49 & \textbf{57} & 62 & \textbf{73} & 54 & \textbf{66} & 48 & \textbf{61} & 71 & 71\\ 
Speckle Noise  & 26 & \textbf{36} & \textbf{67} & 60 & 32 & \textbf{62} & 28 & \textbf{33} & 71 & \textbf{77} \\ 
Gaussian Blur  & 64 & \textbf{67} & 61 & \textbf{75} & 52 & \textbf{64} & 48 & \textbf{62} & \textbf{71} & 69 \\ 
Saturate  & 56 & \textbf{65} & 69 & \textbf{75} & 44 & \textbf{72} & 55 & \textbf{63} & 46 & \textbf{66}\\ 
Gaussian Noise  & 25 & \textbf{31} & \textbf{56} & 51 & 29 & \textbf{58} & 24 & \textbf{28} & 62 & \textbf{73} \\ 
Shot Noise  & 26 & \textbf{30} & \textbf{54} & 49 & 29 & \textbf{63} & 24 & \textbf{28}  & 65 & \textbf{75} \\ 
Impulse Noise  & 32 & \textbf{46} & \textbf{60} & 49 & 37 & \textbf{65} & 29 & \textbf{35}  & 60 & \textbf{72}\\ 
Defocus Blur  & 64 & \textbf{65} & 62 & \textbf{74} & 52 & \textbf{64} & 48 & \textbf{62}  & \textbf{70} & 69 \\ 
Zoom Blur  & 63 & \textbf{68} & 60 & \textbf{72} & 50 & \textbf{60} & 46 & \textbf{62} & 63 & \textbf{64} \\ 
Motion Blur  & 63 & \textbf{68} & 61 & \textbf{73} & 52 & \textbf{63} & 48 & \textbf{63} & \textbf{71} & 68 \\ 
Snow  & \textbf{41} & 38 & 60 & \textbf{69} & 43 & \textbf{62} & \textbf{26} & 25 & 47 & \textbf{62} \\ 
\end{tabular}
\end{center}
\caption{\footnotesize\textit{Results of Acne Lesion Classification with Data Augmentation.} Here we present the accuracies of various classifiers for predicting the lesion type. The \textit{Baseline} and  memory classifier use a ResNet50 backbone with different data augmentation schemes as shown.}
\label{tab:skin-results}
\end{table*}
\begin{figure*}[th]
    \centering
    \includegraphics[height = 4cm, width = 16cm]{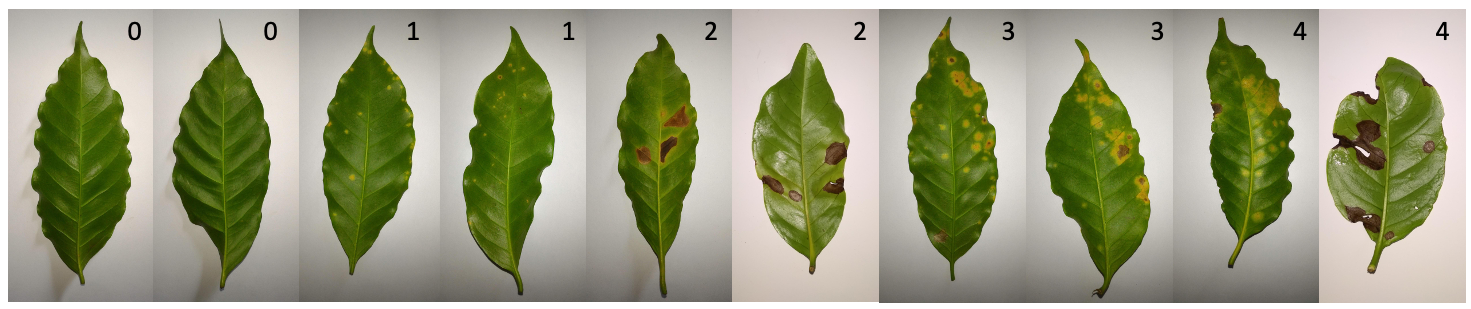}
    \caption{\footnotesize\textit{Case Study: Leaf disease severity.} The figure shows datapoints corresponding to different levels of leaf disease severity. The numbers on each image are the assigned severity labels. }
    \label{fig:leaf_sample}
\end{figure*}

Acne vulgaris, or simply acne, is a common disease that affects approximately $85\%$ of people over their lifespan \citep{metin-gurcan-acne}. The effects of acne and the resulting scars can be psychosocial and possibly psychological, in some cases over the entire lifespan. Distinguishing and counting primary acne lesions, commonly divided into non-inflammatory (comedones, either white or blackheads) and inflammatory ones (pustules, papules, and nodules), is necessary to assess acne severity 
\citep{tan2012}. 
Counting lesions, however, is a very time-consuming task which is usually only performed in clinical trials and required training raters to achieve higher reliability \citep{tan2006}.
The classification task is to assign each primary lesion to one of the three classes commonly used in acne severity scales \citep{acne-grading-scale}: comedones (ClassA), pustules and papules (ClassB), and nodules (ClassC). 
Figure \ref{fig:acne_sample} shows sample lesion crops for the 3 classes.
%
All images were taken from a retrospective deidentified dataset of 63 pediatric patients 
diagnosed with acne, ages 2-21, at the Children's Hospital of Philadelphia
and selected to adequately span the range of acne types encountered in the clinic. 
An expert labeled each image outlining all primary lesions.
%
Individual lesion images were then obtained by cropping out regions containing each lesion, using a binary mask, and resizing them into images of size $300 \times 300$.

A clinician uses features such as the inflammation (redness), lesion size, amount of bulge, among others, when classifying a lesion. These features are usually context-dependent. For instance, the degree of inflammation relies on cues about the skin tone in its immediate neighborhood. 
Thus, the redness of a lesion belonging to the same class can differ across patients. 
Another important cue is the size of the lesion. 
This can depend on the relative size of a lesion compared to other lesions on the same face image. 

%
%

\textbf{Features:} We compute the following three features: 
1) Lesion size $Sz \in \reals$, \ie number of pixels that a lesion occupies in an image, normalized by the median size of lesions over the original face image; 
2) Skin redness $Rd \in \reals$, \ie an indicator of inflammation. We compare the red channel histogram of a lesion with a neighboring skin patch of similar size, as a measure of the degree of redness of the skin. The histogram comparison was implemented using standard computer vision tools \citep{histogram}.  
 %
 Finally, 3) Saturation $Sat \in \{0,1\}$ due to over-exposure. Although not a clinical feature, $Sat$ is an important visual marker of a lesion. Some skin lesions can suffer from over-exposure due to larger than average height as compared to the surrounding face area. This causes light to reflect off of the lesion, making it less detectable. 

\textbf{Experiments:} The dataset consists of a total of $1683$ lesions extracted from the $63$ face images. Out of these, $1153$ lesions from $36$ face images are used for training, and $530$ lesions from $26$ face images for testing. 
To construct the similarity metric, we train a depth-three decision tree $\mathcal{T}$, which uses features $Sz$, $Rd$, and $Sat$ to classify among the three classes - (ClassA, ClassB, ClassC). This gives a binary valued similarity metric, such that $\score(\mathcal{X}_1, \mathcal{X}_2) = 1$ if they are assigned the same class by $\mathcal{T}$. This results in a total of $3$ memories. 

The individual leaf classifiers are trained by transfer learning from a pre-trained deep neural network. We trained the networks with an $SGD$ optimizer for $40$ epochs, with a batch size of $24$, learning rate of $0.01$, and a momentum of $0.9$. When comparing with a data-augmentation scheme, we trained the networks for individual memories after  initializing the network weights with a classifier trained on the whole data augmented training set. To evaluate the robustness of similar techniques, we generated corrupted versions of the face images in the test set  using $15$ different types of image corruptions across $5$ severity levels, as discussed in \cite{hendrycks}. 

\textbf{Comparison with data augmentation methods:} A standard technique to improve robustness of classifiers is to train with an enriched dataset with slightly perturbed inputs. There are several data augmentation techniques in the literature, like AugMix \citep{hendrycks2019augmix}, NoisyMix \citep{noisy_mix_paper}, CutMix\citep{cut_mix}, Manifold Mix \citep{manifold_mix}\footnote{Note that for ManifoldMix comparison, we use their best performing model PreActResNet18 instead}, and Adversarial Training \citep{adversarial_training} among others. 
The results with these methods are presented in Figure \ref{fig:acne_data_aug_comp}, and the detailed table is in the Appendix. We observe that memory classifiers can improve upon these SOTA data-augmentation techniques. We observe an improvement of around $9\%$ across varying severity levels averaged across different perturbation types.

\textbf{Comparison with feature augmentation methods:} 
An alternative way to include features in the training process is to add them the features extracted by the CNN \citep{feature-augment-1, feature-augment-2}. 
Hence, we concatenate the features extracted by a CNN with additional ones we implemented, and train a classifier. We perform our experiments on five candidate architectures for lesion classifiers: ResNet18, ResNet34, ResNet50, VGG11, and VGG16 \cite{He2015,simonyan2014very}. The clean accuracies on the test set are presented in Table \ref{tab:skin-results}. The baselines are trained using the same network architecture as the leaf classifiers for a fair comparison, since our node classifiers can be arbitrary. 
In Table \ref{tab:skin-results}, we observe that memory classifiers are at least as good, or better than, the baseline classifiers. 

\begin{figure}[htb]
\centering
\includegraphics[height = 5cm, width=15cm]{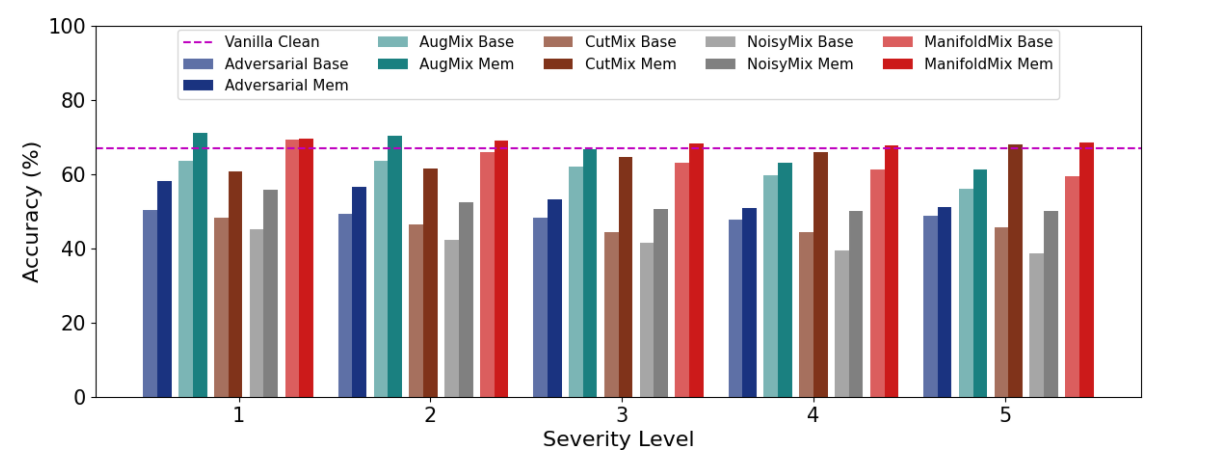}
\caption{\footnotesize Robust accuracy on corrupted acne dataset across five severity levels.}
\label{fig:acne_data_aug_comp}
\end{figure}

\begin{table*}[ht]
\tiny
\begin{center}

\begin{tabular}{ p{1.5cm} p{0.5cm} p{1cm} p{0.5cm} p{1cm}  p{0.5cm} p{1cm} p{0.5cm} p{1cm} p{0.5cm} p{1cm}  }
    \hline
    & \multicolumn{2}{ c }{Adversarial} &  \multicolumn{2}{c}{AugMix} & \multicolumn{2}{c}{CutMix} &  \multicolumn{2}{c}{NoisyMix} & \multicolumn{2}{c}{ManifoldMix} \\
    
    \hline 
    & Baseline & Memory Classifier &  Baseline & Memory Classifier & Baseline & Memory Classifier &  Baseline & Memory Classifier & Baseline & Memory Classifier \\
    \hline 
    Clean Accuracy & 79 & \textbf{80} & 78 & \textbf{79} & 69 & \textbf{75} & 84 & \textbf{87} & \textbf{77} & 75\\
    \hline
    \hline
        Corruption &  \multicolumn{8}{c}{Average Accuracy} \\
    \hline
Fog  & \textbf{64} & 63 & 63 & \textbf{67} & {53} & 39 & 42 & \textbf{66} & 29 & \textbf{39} \\ 
Brightness  & 68 & \textbf{71} & 62 & \textbf{67} & 52 & \textbf{59} & \textbf{69} & 66 & 66 & \textbf{70}\\ 
Contrast  & \textbf{37} & 36 & 37 & \textbf{50} & 25 & 25 & 12 & \textbf{54} & 23 & \textbf{43}\\ 
Elastic  & \textbf{76} & 75 & \textbf{75} & 74 & 70 & \textbf{77} & 83 & \textbf{86} & 77 & 77\\ 
Pixelate  & \textbf{79} & 77 & 76 & \textbf{78} & 70 & \textbf{78} & 83 & \textbf{89} & \textbf{78} & 77  \\ 
JPEG  & 57 & \textbf{61} & 54 & \textbf{59} & 67 & \textbf{70} & 78 & \textbf{82} & \textbf{69} & 68\\ 
Speckle Noise  & 60 & 60 & 60 & \textbf{63} & 68 & \textbf{72} & 80 & \textbf{81} & 66 & \textbf{68} \\ 
Gaussian Blur  & 67 & \textbf{70} & 63 & \textbf{73} & 69 & \textbf{77} & 82 & \textbf{89} & \textbf{79} & 77\\ 
Spatter  & 46 & \textbf{47} & 50 & 50 & 67 & 67 & 65 & \textbf{74} & 47 & \textbf{48} \\ 
Saturate  & 41 & \textbf{49} & 42 & \textbf{50} & \textbf{36} & 33 & 33 & \textbf{40} & 24 & \textbf{34} \\ 
Gaussian Noise  & 50 & 50 & 49 & \textbf{54} & \textbf{68} & 63 & 55 & \textbf{63} & 49 & \textbf{51}\\ 
Shot Noise  & 49 & 49 & 48 & \textbf{55} & 69 & 69 & 61 & \textbf{64} & 50 & \textbf{54}\\ 
Impulse Noise  & 53 & 53 & 52 & \textbf{58} & \textbf{69} & 65 & 57 & \textbf{64} & 50 & 50\\ 
Defocus Blur  & 63 & \textbf{70} & 61 & \textbf{73} & 68 & \textbf{75} & 82 & \textbf{88} & \textbf{78} & 76\\ 
Zoom Blur  & 56 & \textbf{64} & 47 & \textbf{59} & 65 & 65 & 75 & \textbf{82} & \textbf{78} & 71\\ 
Frost  & 29 & \textbf{31} & 31 & 31 & 31 & \textbf{44} & 28 & 28 & \textbf{25} & 24\\ 
Motion Blur  & 69 & \textbf{74} & 64 & \textbf{73} & 68 & \textbf{76} & 82 & \textbf{88} & \textbf{78} & 77 \\ 
Snow  & 37 & \textbf{47} & \textbf{36} & 31 & 21 & \textbf{55} & 32 & \textbf{36} & 26 & \textbf{27}\\ 
\end{tabular}
\end{center}
\caption{\footnotesize\textit{Results of Leaf Severity Classification with Data Augmentation.} We present the accuracies of various classifiers for predicting leaf disease severity. The \textit{Baseline} and memory classifier use a ResNet50 backbone with various data augmentation schemes as shown.}
\label{tab:leaf-severity-data-aug-results}
\end{table*}

\subsection{Case Study: Leaf Disease Severity}

Machine learning has the ability to transform the agricultural industry through crop yield prediction, intelligent spraying, and disease diagnosis. The task of identifying plant diseases is a promising problem for machine learning. 
In this case study, we focus on a dataset of coffee leaves affected by multiple biotic stress factors, such as leaf miners, rust, brown leaf spots, and cercospora leaf spots. 
Our dataset contains 1747 images of arabica coffee leaves, including healthy and diseased leaves. We use $80\%$ of the dataset for training, and the remainder for test. We refer the reader to \citet{coffee_dataset} for more details. 
In Figure \ref{fig:leaf_sample}, we show a sample of leaves with varying severity levels, depending on the degree and type of the affected area. This is available in the dataset as a severity score $SV \in \{0, \dots, 4\}$, where a higher score denotes more severity. 
The classification task is to predict the severity level from the input image. 

\textbf{Features:} As a pre-processing step, we partition a given image into different segments via \cite{Felzenszwalb2004} 
and perform a color value thresholding in HSV color space to obtain the green, brown, and discolored regions. 
We over-approximate the leaf region by constructing a convex hull of the green pixels. This ensures that any \emph{holes} created by discoloration are included as an area of the leaf. We then define functions $F_d :\mathcal{X} \rightarrow [0,1]$ and $F_b : \mathcal{X} \rightarrow [0,1]$ as the number of pixels discovered as discolored or brown as a fraction of the green pixels. Figure 4 in Appendix, shows sample masks overlaid on the original image.
Such features can inform health of a leaf: a plant with mostly green leaves without discolored or brown areas is likely to be healthy.
A decision tree $\mathcal{T}: [0,1]^2 \rightarrow \{0, \dots, 4\}$ based on these features classifies the image into one of the five severity levels, so that
$\mathcal{T} ( F_d(x), F_b(x) ) \in SV$.
Given the label computed by $\mathcal{T}$, two leaf images with the same predicted class are assigned a $\score$ score of $1$, and $0$ otherwise. 

\textbf{Experiments:} We train leaf classifiers and baseline classifiers as discussed in the previous case study. It takes $\sim 30$ minutes for training our memory classifier on a single NVIDIA Quadro RTX 6000 GPU. In \cite{coffee_dataset}, the authors perform transfer learning on a pre-trained ResNet50 
architecture to predict the severity score. 
Here we train the networks on the original image, instead of cropping out the leaf area as a preprocessing step as in \cite{coffee_dataset}. This makes the problem more challenging, and causes a $4\%$ drop in test accuracy. We study the robustness of different neural network architectures and data augmentation methods similarly to the previous case study. We subject the leaf images to $18$ different corruptions across $5$ different severity levels.

\begin{figure}[htb]
\centering
\includegraphics[height = 5cm, width=15cm]{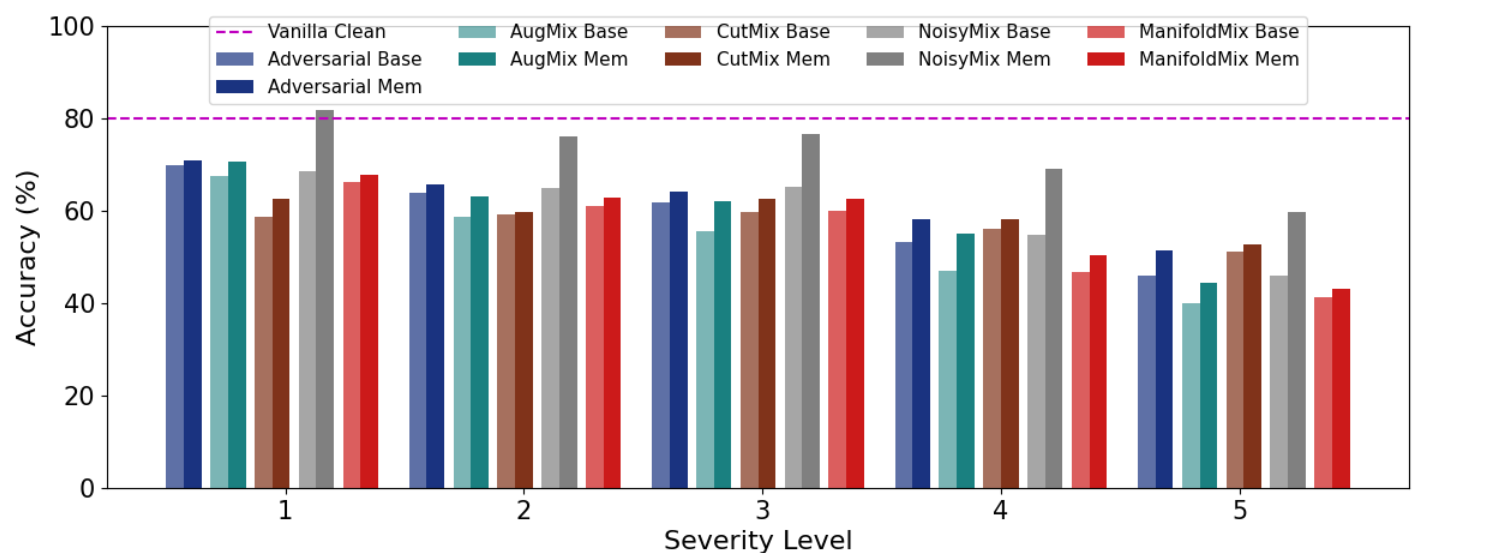}
\caption{\footnotesize Robust accuracy on corrupted leaf dataset across five severity levels.}
\label{fig:leaf_data_aug_comp}
\end{figure}

\textbf{Comparison with data augmentation methods:} As mentioned in the previous case study, we use data augmentation for improving the accuracy, using NoisyMix, AugMix, CutMix, ManifoldMix and Adversarial Training. The results are presented in Table \ref{tab:leaf-severity-data-aug-results}. 
We observe that across different data augmentation schemes, memory classifiers achieve consistently better accuracy under different degrees of severity in the image noise. We plot the results in Figure \ref{fig:leaf_data_aug_comp}.

\textbf{Comparison with feature augmentation methods:} We augment the features extracted by a CNN with the features discussed above, and train a baseline classifier using them. 
We repeat this experiment on $5$ different CNN architectures for both the leaf classifiers and the standalone CNN as the baseline (Table \ref{tab:leaf-severity-feat-aug-results}). The clean accuracies of the memory classifiers and the baseline network are  comparable.
However, memory classifiers are consistently more robust than the  simple feature augmented counterparts.

\section{Related Work}
\label{sec:related_work}
An early attempt at case-based reasoning---which is distantly related to our approach of clustering to a small set of memories---can be found in \cite{prototype_paper_1}. There has been a growing interest in using these ideas  for image classification. The motivations range from interpretability  \citep{prototype_paper_2} to solving few shot learning \citep{prototype_paper_3}, and even one shot learning \citep{prototype_paper_4}.  Our work is related, but substantially different due to our reliance on  memories; this leads to further differences in implementation and learning guarantees. Another closely related idea is of \emph{coresets} \eg \citet{coresets}, \etc, which uses a subset of representative data samples for training. 
This can improve training time, while usually achieving similar performance as using the full data. In contrast, our clustering step via the similarity function approach aims to improve performance and robustness.  Our workflow is structurally similar to the idea of Mixtures-of-Experts \citep{moe_paper} (MoE) models but is conceptually different. A challenge in MoE models is to define their regions of activity. The novelty in our work is that these regions are defined via the memories in conjunction with the distance metric.
Other prior work includes efforts to de-bias DNNs by designing training sets \citep{de-bias-neural-net}; understanding how shape-bias disagreement occurs in DNNs \citep{understand-shape-bias}; and using Gabor filters to simulate the visual cortex \citep{gabor-filter}.  Other conceptually similar work includes \citet{snorkel, feature-augment-1, feature-augment-2}.




\section{Discussion}
\label{sec:conclusions}

We show that using robust features to compute a similarity metric offers a viable path towards robustness. 
Clustering the dataset using the robust features allows the CNN to improve on inductive biases present in each cluster. Consider the example of the color dataset discussed in Section \ref{sec:introduction}. The features extract the color of the patch in the image, which decides the similarity. For the cluster belonging to a memory of the green patch, the trained CNN is more likely to label an image as a green patch even after corruption, simply because of the distribution of the training data. This explains some of the robustness we observe.  Our experiments suggest that this robustness also extends to more complex datasets. Construction of such additional robust features will be explored in the future.
\bibliographystyle{plainnat}
\bibliography{references}

\appendix

\section{Appendix}

\label{sec:memory_classifier_theory}

\subsection{Generalization properties}

\textbf{Definitions and notations.} 
We will use a few definitions and notations: for a positive integer $q$, we denote $[q]:=\{1,2,\ldots,q\}$.  We denote by $I(A)$ the indicator of the event $A$, which equals to unity (\ie ``$1$") if event $A$ is true, and equals to zero if event $A$ is false. 

Now we provide the proof of the data-dependent generalization bound \eqref{eq:generalization_bound} for \mcs. 
We let the training and testing samples be independently and identically distributed (iid), sampled
from a fixed unknown distribution $D$ over the space $\mathcal{X} \times\{-1, +1\}$. 
Let $\mathcal{S}'$ be the hypothesis class that corresponds to using $s$ with the thresholds $b_k$, $k\in [q]$, as described in the text.
Let $\mathcal{S}_k = \{s(\cdot,k): s\in \mathcal{S}\}$, and define  $\mathcal{S}'_k$ similarly.
Consider a set of iid random variables $\varepsilon_i$, $i\in [n]$, distributed uniformly over $\{-1,1\}$.
For a hypothesis class $H$, let the empirical Rademacher complexity of the hypothesis space $H$ over the training set $T$ be
$\smash{\hat{\mathfrak{R}}_T(H)} = n^{-1} \mathbb{E}_{\ep_1,\ldots, \ep_n}\sup_{h\in H}\sum_{i=1}^n \varepsilon_i h(x_i)$ \citep{koltchinskii2002empirical,bartlett2002rademacher,shalev-shwartz_ben-david_2014}. 

A key insight is that memory classifiers can be thought of as cascade classifiers \citep{deep_cascade}.
From their Theorem 1 (with $p=1$), we have 
\begin{equation}\label{ib}
    R(F) \leq \hat{R}_T(F) + 
    \sum_{k=1}^{q}\min\left(\frac{n_k^+}{n}, 4 [\hat{\mathfrak{R}}_T(H_k) + \hat{\mathfrak{R}}_T(\mathcal{S}'_k)]\right) + C(n,q+1,\delta).
\end{equation}
Thus, 
it remains to bound the Rademacher complexity of the selector function class $\mathcal{S}'_k$.

Recall that $M = \{ m_1, \dots, m_q \}$ is the set of memories.
Then, for an input $x\in \mathcal{X}$, and a memory $m_k$ with $k\in [q]$, the selector function $s$ chooses the memories most similar to $x$ from the set $M$ 
by computing: 
\begin{equation*}
    s(x,k) = I\left(\s(x,m_k) \geq 
    \max(\{\s(x, m_j), j \in q, j \neq k\}, b_k)\right).
\end{equation*}
If $b_k$ is the larger above, then some hypothesis in $H_{q+1}$ is chosen.
Thus, we have $\mathcal{S}_k'\subset \mathcal{S}_k \cup H_{q+1}$.
Since $\mathcal{S}_k$ consists of 0-1 valued classifiers, and $H_{q+1}$ consists of a constant classifier (thus its empirical Rademacher complexity is of order $O(1/n^{1/2})$, we find, for some constant $\kappa>0$,
\begin{equation}\label{sum}
    \begin{aligned}
    \hat{\mathfrak{R}}_T(\mathcal{S}'_k) \leq \hat{\mathfrak{R}}_T(\mathcal{S}_k) + \kappa/n^{1/2}.
    \end{aligned}
\end{equation}

The number of hypotheses in $\mathcal{S}_k$ that can be formed based on $n$ datapoints is at most $N_{n,q} = n \cdot {n-1 \choose q-1}$.
This is because we can choose the first memory $m_k$ as any of the $n$ datapoints (or the virtual memory), and then we can choose the remaining $q$ memories  as any subset of the remaining $n-1$ datapoints. Each such choice determines at most one distinct function $s(\cdot,k)$, which gives the desired bound.

Let $A = \{a_s:s(\cdot,k)\in\mathcal{S}_k\}$ be the set of binary vectors indexed by $s$, for each $s$ containing the indices of the datapoints such that 
$s(x_i, k) = 1$. Thus $a_s \in \{ 0,1\}^n$, and $a_{s,i} = 1$ iff $s(x_i, k) = 1$. We have $\| a_s\|_2 \leq n^{1/2}$, because $a_s \in \{ 0,1\}^n$. 

Let $\bar{a}$ be the mean of all $a_s$, $s\in\mathcal{S}$. The Massart Lemma (see, e.g., Lemma 26.8 in \cite{shalev-shwartz_ben-david_2014}) 
shows that 
$$\hat{\mathfrak{R}}_T (\mathcal{S}_k) \leq \max_{a_s \in A} || a_s - \bar{a}|| \frac{\log(N_{n,q})}{n}.$$
Now, $\max_{a_s \in A} || a_s - \bar{a} || \leq \max_{a_s \in A}||a_s|| \leq n^{1/2}$.
Moreover,
\begin{align*}
&\log N_{n,q} = \log\left(n \cdot  {n-1 \choose q-1} \right) 
            \leq \log\left(n \cdot \left( e \frac{n-1}{q-1} \right)^{q-1}\right) \\
&\leq \log(n) + (q-1)\left[ 1 + \log{\frac{n-1}{q-1}} \right] \leq q \cdot(1 + \log n).
\end{align*}
In the last line we used the somewhat crude bound $\log{\frac{n-1}{q-1}} \le \log n$. We obtain
\begin{equation}
\label{eq:memory_bound}
    \hat{\mathfrak{R}}_T(\mathcal{S}_k) \leq \frac{q \cdot (1 + \log n)}{n^{1/2}}.
\end{equation}
Bounding the minimum in \eqref{ib} by the Rademacher complexity term,
using \eqref{sum} and \eqref{eq:memory_bound},
we find
\begin{equation*}
\begin{aligned}
    R(F) & \leq \hat{R}_T(F) + 
     4q\left[\frac{q (1 + \log n)}{n^{1/2}}+\max_{k=1}^{q} \hat{\mathfrak{R}}_T(H_k) 
     + \kappa/n^{1/2} \right] + C(n,q,\delta).
\end{aligned}
\end{equation*}
This finishes the proof.

\subsection{Computing Features for the Leaf}
\label{sec:features_leaf}
\begin{figure*}[h]
\centering
\includegraphics[height = 10cm, width = 13cm]{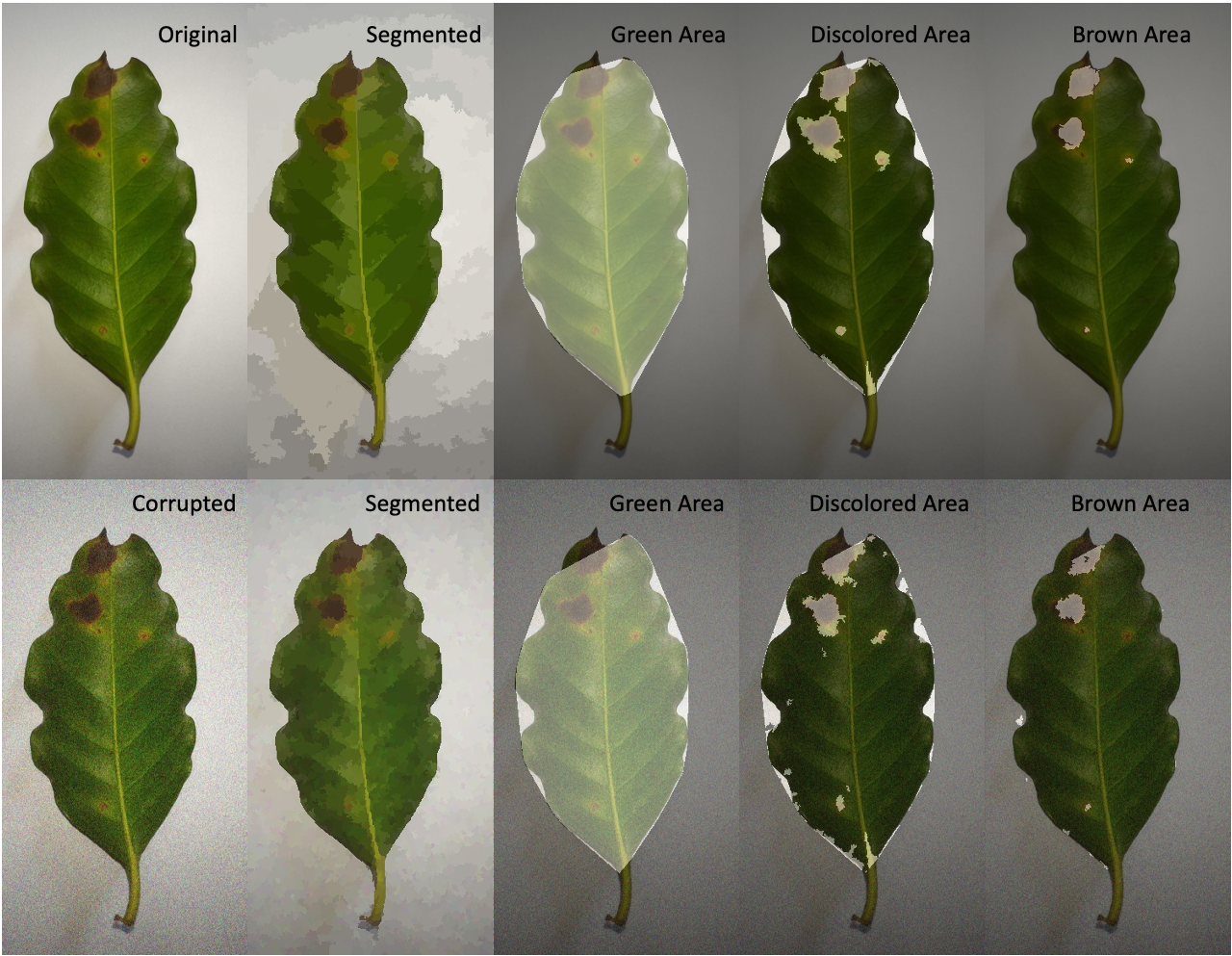}
\caption{The two rows compare feature extraction in a clean image vs in a corrupted image. The starting images are shown on the left. The corruption is obtained by adding Gaussian noise with an intensity of $3$; see \cite{hendrycks}, for details.
One can observe that adding noise does not completely impair the feature extraction, and thus helps in improving the robustness of our classification process.
    }
    \label{fig:leaf_processing}
\end{figure*}
Figure \ref{fig:leaf_processing} demonstrates the different masks overlaid on the original image. The number of pixels discovered as discolored/brown as a fraction of the green pixels are returned by the functions $F_d$ and $F_b$, respectively. We also demonstrate how these features are robust to natural perturbations.

\subsection{Ablation on Different CNN Architectures}
We present the results for different CNN architectures in Table \ref{tab:acne_severity_diff_arch_results}, and Table  \ref{tab:leaf-severity-diff-arch-results} for the leaf severity and lesion classification datasets. The clean test accuracies obtained by the memory classifiers are comparable to those of standard neural nets, 
across the datasets and network architectures. The performances are in most cases comparable or even better than for the original network.

\begin{table*}[ht]
\tiny
\begin{center}
\begin{tabular}{ p{1.5cm} p{0.5cm} p{1cm} p{0.5cm} p{1cm}  p{0.5cm} p{1cm} p{0.5cm} p{1cm} p{0.5cm} p{1cm}  }

    \hline
    & \multicolumn{2}{ c }{Resnet18} &  \multicolumn{2}{c}{Resnet34} & \multicolumn{2}{c}{Resnet50} &  \multicolumn{2}{c}{VGG11}
    &  \multicolumn{2}{c}{VGG16}\\
    
    \hline 
    & Baseline & Memory Classifier &  Baseline & Memory Classifier& Baseline & Memory Classifier &  Baseline & Memory Classifier &  Baseline & Memory Classifier\\
    \hline 
    Clean Accuracy & 78 & 78 & 68 & \textbf{78} & 68 & \textbf{77} & 75 & \textbf{78} & 73 & \textbf{78} \\
    \hline
    \hline
        Corruption &  \multicolumn{10}{c}{Average Accuracy} \\
    \hline
Brightness  & 66 & \textbf{74} & 70 & \textbf{73} & 71 & \textbf{72} & 72 & \textbf{73} & 68 & \textbf{73} \\ 
Contrast  & 66 & \textbf{77} & 64 & \textbf{77} & 64 & \textbf{75} & 64 & \textbf{77} & 64 & \textbf{77} \\ 
Elastic  & 63 & \textbf{78} & 65 & \textbf{75} & 70 & \textbf{77} & 68 & \textbf{77} & 67 & \textbf{78} \\ 
Pixelate  & 64 & \textbf{75} & 65 & \textbf{73} & 64 & \textbf{73} & 65 & \textbf{78} & 64 & \textbf{77} \\ 
JPEG  & 64 & \textbf{76} & 64 & \textbf{73} & 64 & \textbf{71} & 67 & \textbf{78} & 67 & \textbf{78} \\ 
Speckle Noise  & 62 & \textbf{78} & 75 & \textbf{77} & 74 & \textbf{77} & 70 & \textbf{77} & 76 & \textbf{77} \\ 
Gaussian Blur  & 63 & \textbf{77} & 64 & \textbf{77} & 63 & \textbf{70} & 64 & \textbf{77} & 64 & \textbf{77} \\ 
Saturate  & 68 & \textbf{76} & 70 & \textbf{76} & 65 & \textbf{76} & 70 & \textbf{77} & 68 & \textbf{78} \\ 
Gaussian Noise  & 52 & \textbf{77} & 72 & \textbf{77} & 70 & \textbf{77} & 68 & \textbf{77} & 74 & \textbf{77} \\ 
Shot Noise  & 55 & \textbf{78} & 72 & \textbf{77} & 67 & \textbf{77} & 69 & \textbf{77} & 77 & 77 \\ 
Impulse Noise  & 53 & \textbf{77} & 64 & \textbf{76} & 75 & \textbf{76} & 70 & \textbf{76} & \textbf{78} & 76 \\ 
Defocus Blur  & 63 & \textbf{76} & 63 & \textbf{77} & 63 & \textbf{69} & 64 & \textbf{77} & 63 & \textbf{77} \\ 
Zoom Blur  & 63 & \textbf{76} & 65 & \textbf{75} & 65 & \textbf{75} & 65 & \textbf{76} & 64 & \textbf{76} \\ 
Motion Blur  & 63 & \textbf{76} & 64 & \textbf{76} & 65 & \textbf{74} & 66 & \textbf{76} & 65 & \textbf{77} \\ 
Snow  & 72 & \textbf{77} & 75 & \textbf{76} & 72 & \textbf{75} & \textbf{77} & 72 & 72 & \textbf{75} \\ 
\end{tabular}
\end{center}
\caption{\textit{Acne Lesion Classification Results (Network Ablation).} For a given network architecture, we compare the evaluation on a standalone neural network classifier and the same classifier used in a memory classifier setting. The memories were formed using features corresponding to size, redness, and saturation level of a skin-lesion. We report the average accuracy (in percentage) across $5$ severity levels for different types of image corruption. For a given architecture and corruption, we highlight the ones with better accuracy. Note that, in dermatology, above $75\%$ are considered acceptable accuracies also for human raters \cite{tan2006}.}
\label{tab:acne_severity_diff_arch_results}
\end{table*}
\begin{table*}[t]
\begin{center}
\tiny
\begin{tabular}{ p{1.5cm} p{0.5cm} p{1cm} p{0.5cm} p{1cm}  p{0.5cm} p{1cm} p{0.5cm} p{1cm} p{0.5cm} p{1cm}  }

\hline
Backbone                                                 & \multicolumn{2}{c}{ResNet 18}                                             & \multicolumn{2}{c}{ResNet 34}                                             & \multicolumn{2}{c}{ResNet 50}                                             & \multicolumn{2}{c}{VGG11}                                                 & \multicolumn{2}{c}{VGG16}                                                 \\ \hline
                                                         & Baseline    & \begin{tabular}[c]{@{}c@{}}Memory \\Classifier\end{tabular} & Baseline    & \begin{tabular}[c]{@{}c@{}}Memory \\Classifier\end{tabular} & Baseline    & \begin{tabular}[c]{@{}c@{}}Memory \\Classifier\end{tabular} & Baseline    & \begin{tabular}[c]{@{}c@{}}Memory \\Classifier\end{tabular} & Baseline    & \begin{tabular}[c]{@{}c@{}}Memory \\Classifier\end{tabular} \\ \hline
\begin{tabular}[c]{@{}c@{}}Clean Accuracy\end{tabular} & 75          & \textbf{78}                                                 & \textbf{77} & 76                                                          & \textbf{80} & 77                                                          & 71 & 71   & 63 & 69                                               \\ \hline
\hline
Corruption                                               & \multicolumn{10}{c}{Average Accuracy}                                                                                                                                                                                                                                                                                                                                                     \\ \hline
Fog                                                      & \textbf{60} & 55                                                          & \textbf{63} & 52                                                          & \textbf{64} & 62                                                          & 43          & \textbf{49}                                                 & 42          & \textbf{44}                                                 \\
Brightness                                               & 60          & \textbf{61}                                                 & 68          & \textbf{73}                                                 & \textbf{69} & 67                                                          & 66 & 66                                               & \textbf{67} & 66                                                          \\
Contrast                                                 & 37          & \textbf{42}                                                 & 33          & 30                                                          & 34          & \textbf{45}                                                 & 23          & \textbf{39}                                                 & 24          & \textbf{41}                                                 \\
Elastic                                                  & 73          & \textbf{75}                                                 & 75          & \textbf{76}                                                 & 74 & 74                                                 & 67          & \textbf{68}                                                 & 62          & \textbf{69}                                                 \\
Pixelate                                                 & 75          & \textbf{76}                                                 & \textbf{77} & \textbf{77}                                                 & \textbf{77} & 74                                                          & 69          & \textbf{71}                                                 & 63          & \textbf{70}                                                 \\
JPEG                                                     & 61          & \textbf{66}                                                 & 65          & \textbf{69}                                                 & 54          & \textbf{56}                                                 & 60          & \textbf{65}                                                 & 58          & \textbf{63}                                                 \\
Speckle Noise                                            & 66          & \textbf{69}                                                 & 51          & \textbf{68}                                                 & \textbf{62} & 61                                                          & 55          & \textbf{61}                                                 & 49          & \textbf{65}                                                 \\
Gaussian Blur                                            & 71          & \textbf{73}                                                 & 69          & \textbf{75}                                                 & \textbf{63} & 61                                                          & 61          & \textbf{64}                                                 & 64          & \textbf{67}                                                 \\
Spatter                                                  & 49          & \textbf{62}                                                 & 65 & 65                                                 & 49 & 49                                                 & 50          & \textbf{51}                                                 & 46 & \textbf{58}                                                 \\
Saturate                                                 & 44          & \textbf{57}                                                 & 52          & \textbf{59}                                                 & 41          & \textbf{55}                                                 & 52          & \textbf{55}                                                 & 44          & \textbf{52}                                                 \\
Gaussian Noise                                           & 57          & \textbf{62}                                                 & 49          & \textbf{65}                                                 & 50          & \textbf{54}                                                 & 46          & \textbf{53}                                                 & 43          & \textbf{61}                                                 \\
Shot Noise                                               & 58          & \textbf{63}                                                 & 48          & \textbf{66}                                                 & 51          & \textbf{55}                                                 & 46          & \textbf{53}                                                 & 45          & \textbf{63}                                                 \\
Impulse Noise                                            & 56          & \textbf{60}                                                 & 45          & \textbf{64}                                                 & 54          & \textbf{58}                                                 & 45          & \textbf{50}                                                 & 45          & \textbf{61}                                                 \\
Defocus Blur                                             & 70          & \textbf{73}                                                 & 67          & \textbf{74}                                                 & 60 & 60                                                 & 60          & \textbf{62}                                                 & 63          & \textbf{65}                                                 \\
Zoom Blur                                                & 65          & \textbf{67}                                                 & \textbf{63} & 62                                                          & 52          & \textbf{58}                                                 & 56          & \textbf{53}                                                 & \textbf{55} & 50                                                          \\
Frost                                                    & 31          & \textbf{41}                                                 & 41          & \textbf{42}                                                 & 29          & \textbf{43}                                                 & \textbf{49} & 45                                                          & \textbf{41} & 32                                                          \\
Motion Blur                                              & 74 & 74                                                 & 74          & \textbf{75}                                                 & 64          & \textbf{65}                                                 & 62          & \textbf{64}                                                 & 63          & \textbf{65}                                                 \\
Snow                                                     & 42          & \textbf{53}                                                 & 48          & \textbf{59}                                                 & 38          & \textbf{53}                                                 & 48          & \textbf{49}                                                 & 42          & \textbf{46}                                                
\end{tabular}
\end{center}
\caption{\textit{Leaf Disease Severity Results (Network Ablation).} For a given network architecture, we compare the evaluation on a standalone DNN classifier and the same network used in a memory classifier setting. The memories were formed using features corresponding to different proportions of discoloration in the leaf. We report the average accuracy (in percentage) across $5$ severity levels for different types of image corruption. For a given architecture and corruption, we highlight the ones with better accuracy.}
\label{tab:leaf-severity-diff-arch-results}
\end{table*}

\subsection{Augmented Features Results}

An alternate way to add features to the ones extracted by a DNN is by \emph{feature augmentation}, including them in the penultimate layer. The intuition is that the final layer might be able learn how to incorporate them with features extracted from the images. 
The results for this approach are presented in Tables \ref{tab:aug-skin-results} and \ref{tab:leaf-severity-feat-aug-results}, for the skin and leaf datasets, respectively.

\label{sec:feature-augment-results}
\begin{table*}[ht]
\begin{center}
\tiny
\begin{tabular}{ccccccccccc}
\hline
Backbone                                                 & \multicolumn{2}{c}{ResNet 18}                                          & \multicolumn{2}{c}{ResNet 34}                                          & \multicolumn{2}{c}{ResNet 50}                                          & \multicolumn{2}{c}{VGG11}                                              & \multicolumn{2}{c}{VGG16}                                              \\ \hline
                                                         & Baseline & \begin{tabular}[c]{@{}c@{}}Memory\\ Classifier\end{tabular} & Baseline & \begin{tabular}[c]{@{}c@{}}Memory\\ Classifier\end{tabular} & Baseline & \begin{tabular}[c]{@{}c@{}}Memory\\ Classifier\end{tabular} & Baseline & \begin{tabular}[c]{@{}c@{}}Memory\\ Classifier\end{tabular} & Baseline & \begin{tabular}[c]{@{}c@{}}Memory\\ Classifier\end{tabular} \\ \hline
\begin{tabular}[c]{@{}c@{}}Clean\\ Accuracy\end{tabular} & 66       & \textbf{78}                                                 & 70       & \textbf{78}                                                 & 67       & 77                                                 & 77       & \textbf{78}                                                 & 77       & \textbf{78}                                                 \\ \hline \hline
Corruption                                               & \multicolumn{10}{c}{Average Accuracy}                                                                                                                                                                                                                                                                                                                                      \\ \hline
Brightness                                               & 55       & \textbf{74}                                                 & 63       & \textbf{73}                                                 & 50       & \textbf{72}                                                 & 54       & \textbf{73}                                                 & 54       & \textbf{73}                                                 \\
Contrast                                                 & 62       & \textbf{77}                                                 & 65       & \textbf{77}                                                 & 64       & \textbf{75}                                                 & 72       & \textbf{77}                                                 & 72       & \textbf{77}                                                 \\
Elastic                                                  & 55       & \textbf{78}                                                 & 61       & \textbf{75}                                                 & 60       & \textbf{77}                                                 & 64       & \textbf{77}                                                 & 64       & \textbf{78}                                                 \\
Pixelate                                                 & 63       & \textbf{75}                                                 & 65       & \textbf{73}                                                 & 60       & \textbf{73}                                                 & 65       & \textbf{78}                                                 & 65       & \textbf{77}                                                 \\
JPEG                                                     & 64       & \textbf{76}                                                 & 65       & \textbf{73}                                                 & 63       & \textbf{71}                                                 & 58       & \textbf{78}                                                 & 58       & \textbf{78}                                                 \\
Speckle Noise                                            & 43       & \textbf{78}                                                 & 73       & \textbf{77}                                                 & 32       & \textbf{77}                                                 & 74       & \textbf{77}                                                 & 74       & \textbf{77}                                                 \\
Gaussian Blur                                            & 62       & \textbf{77}                                                 & 66       & \textbf{77}                                                 & 63       & \textbf{70}                                                 & 56       & \textbf{77}                                                 & 56       & \textbf{77}                                                 \\
Saturate                                                 & 64       & \textbf{76}                                                 & 62       & \textbf{76}                                                 & 64       & \textbf{76}                                                 & 53       & \textbf{77}                                                 & 53       & \textbf{78}                                                 \\
Gaussian Noise                                           & 40       & \textbf{77}                                                 & 73       & \textbf{77}                                                 & 39       & \textbf{77}                                                 & 74       & \textbf{77}                                                 & 74       & \textbf{77}                                                 \\
Shot Noise                                               & 38       & \textbf{78}                                                 & 70       & \textbf{77}                                                 & 35       & \textbf{77}                                                 & 75       & \textbf{77}                                                 & 75       & \textbf{77}                                                 \\
Impulse Noise                                            & 43       & \textbf{77}                                                 & 59       & \textbf{76}                                                 & 44       & \textbf{76}                                                 & 71       & \textbf{76}                                                 & 71       & \textbf{76}                                                 \\
Defocus Blur                                             & 63       & \textbf{76}                                                 & 66       & \textbf{77}                                                 & 64       & \textbf{69}                                                 & 56       & \textbf{77}                                                 & 56       & \textbf{77}                                                 \\
Zoom Blur                                                & 60       & \textbf{76}                                                 & 64       & \textbf{75}                                                 & 64       & \textbf{75}                                                 & 57       & \textbf{76}                                                 & 57       & \textbf{76}                                                 \\
Motion Blur                                              & 63       & \textbf{77}                                                 & 66       & \textbf{76}                                                 & 63       & \textbf{74}                                                 & 59       & \textbf{76}                                                 & 59       & \textbf{77}                                                
\end{tabular}
\end{center}
\caption{\textit{Lesion Classification Results (Feature Augmentation).} The baseline network is a standard DNN augmented with the special features we extract. We compare this with memory classifiers with the same network structure, but without the extra features. We report the average accuracy (as a percentage) across five severity levels, for different types of image corruption. For a given architecture and corruption, we highlight the ones with better accuracy. }
\label{tab:aug-skin-results}
\end{table*}
\begin{table*}[ht]
\begin{center}
\tiny
\begin{tabular}{ p{1.5cm} p{0.5cm} p{1cm} p{0.5cm} p{1cm}  p{0.5cm} p{1cm} p{0.5cm} p{1cm} p{0.5cm} p{1cm}  }
\hline
Backbone                                                 & \multicolumn{2}{c}{ResNet 18}                                             & \multicolumn{2}{c}{ResNet 34}                                             & \multicolumn{2}{c}{ResNet 50}                                             & \multicolumn{2}{c}{VGG11}                                                 & \multicolumn{2}{c}{VGG16}                                                 \\ \hline
                                                         & Baseline    & \begin{tabular}[c]{@{}c@{}}Memory\\ Classifier\end{tabular} & Baseline    & \begin{tabular}[c]{@{}c@{}}Memory\\ Classifier\end{tabular} & Baseline    & \begin{tabular}[c]{@{}c@{}}Memory\\ Classifier\end{tabular} & Baseline    & \begin{tabular}[c]{@{}c@{}}Memory\\ Classifier\end{tabular} & Baseline    & \begin{tabular}[c]{@{}c@{}}Memory\\ Classifier\end{tabular} \\ \hline
\begin{tabular}[c]{@{}c@{}}Clean\\ Accuracy\end{tabular} & 74          & \textbf{78}                                                 & \textbf{77} & 76                                                          & \textbf{80} & 77                                                          & 67          & \textbf{71}                                                 & 67          & \textbf{69}                                                 \\ \hline \hline
Corruption                                               & \multicolumn{10}{c}{Average Accuracy}                                                                                                                                                                                                                                                                                                                                                     \\ \hline
Fog                                                      & \textbf{59} & 55                                                          & \textbf{61} & 52                                                          & \textbf{64} & 62                                                          & 41          & \textbf{49}                                                 & 41          & \textbf{44}                                                 \\
Brightness                                               & 59          & \textbf{61}                                                 & 66          & \textbf{73}                                                 & \textbf{69} & 67                                                          & 62          & \textbf{66}                                                 & 62          & \textbf{66}                                                 \\
Contrast                                                 & 34          & \textbf{42}                                                 & 30 & 30                                                & 33          & \textbf{45}                                                 & 21          & \textbf{39}                                                 & 21          & \textbf{41}                                                 \\
Elastic                                                  & 74          & \textbf{75}                                                 & 76 & 76                                                & \textbf{75} & 74                                                          & 63          & \textbf{68}                                                 & 63          & \textbf{69}                                                 \\
Pixelate                                                 & 75          & \textbf{76}                                                 & 77 & 77                                                 & \textbf{77} & 74                                                          & 66          & \textbf{71}                                                 & 66          & \textbf{70}                                                 \\
JPEG                                                     & 62          & \textbf{66}                                                 & 62          & \textbf{69}                                                 & 53          & \textbf{56}                                                 & 56          & \textbf{65}                                                 & 56          & \textbf{63}                                                 \\
Speckle Noise                                            & 66          & \textbf{69}                                                 & 52          & \textbf{68}                                                 & 60          & \textbf{61}                                                 & 51          & \textbf{61}                                                 & 51          & \textbf{65}                                                 \\
Gaussian Blur                                            & 70          & \textbf{73}                                                 & 67          & \textbf{75}                                                 & \textbf{64} & 61                                                          & 59          & \textbf{64}                                                 & 59          & \textbf{67}                                                 \\
Spatter                                                  & 51          & \textbf{62}                                                 & 62          & \textbf{65}                                                 & \textbf{51} & 49                                                          & 40          & \textbf{51}                                                 & 40          & \textbf{58}                                                 \\
Saturate                                                 & 41          & \textbf{57}                                                 & 50          & \textbf{59}                                                 & 42          & \textbf{55}                                                 & 51          & \textbf{55}                                                 & 51          & \textbf{52}                                                 \\
Gaussian Noise                                           & 56          & \textbf{62}                                                 & 49          & \textbf{65}                                                 & 50          & \textbf{54}                                                 & 42          & \textbf{53}                                                 & 42          & \textbf{61}                                                 \\
Shot Noise                                               & 58          & \textbf{63}                                                 & 48          & \textbf{66}                                                 & 50          & \textbf{55}                                                 & 42          & \textbf{53}                                                 & 42          & \textbf{63}                                                 \\
Impulse Noise                                            & 56          & \textbf{60}                                                 & 46          & \textbf{64}                                                 & 53          & \textbf{58}                                                 & 41          & \textbf{50}                                                 & 41          & \textbf{61}                                                 \\
Defocus Blur                                             & 67          & \textbf{73}                                                 & 64          & \textbf{74}                                                 & \textbf{62} & 60                                                          & 57          & \textbf{62}                                                 & 57          & \textbf{65}                                                 \\
Zoom Blur                                                & 63          & \textbf{67}                                                 & \textbf{65} & 62                                                          & 51          & \textbf{58}                                                 & \textbf{55} & 53                                                          & \textbf{55} & 50                                                          \\
Blur                                                     & 32          & \textbf{41}                                                 & 40          & \textbf{42}                                                 & 30          & \textbf{43}                                                 & 42          & \textbf{45}                                                 & \textbf{42} & 32                                                          \\
Motion Blur                                              & 71          & \textbf{74}                                                 & 72          & \textbf{75}                                                 & \textbf{66} & 65                                                          & 60          & \textbf{64}                                                 & 60          & \textbf{65}                                                 \\
Snow                                                     & 39          & \textbf{53}                                                 & 46          & \textbf{59}                                                 & 37          & \textbf{53}                                                 & 36          & \textbf{49}                                                 & 36          & \textbf{46}                                                
\end{tabular}
\end{center}
\caption{\textit{Leaf Disease Severity Results (Feature Augmentation).} We follow the same protocol as in Table \ref{tab:aug-skin-results}.}
\label{tab:leaf-severity-feat-aug-results}
\end{table*}

\subsection{Color Dataset Details}
\label{sec:color-dataset-results}

We generate the color dataset as described in Section \ref{sec:introduction}, with $L = 500$ and $w = L/10$. The three classes are generated using the parameter $p$ set to $[(255,0,0), (0,255,0), (0, 0,255)]$. This generates images belonging to three color classes. The dataset has $3000$ training and $300$ test images.

We train the following baselines: 
\begin{enumerate}
    \item ResNet50: The network is trained using transfer learning from a pre-trained model. We use an SGD optimizer to train for $50$ epochs, with a learning rate of $0.01$, weight decay $0.0005$, batch size of $48$ and momentum of $0.9$. The trained network has $100\%$ accuracy on the clean test images.
    
    \item AugMix: The training uses a similar ResNet50 backbone network, with similar hyper-parameters and training algorithm, as discussed in \citep{hendrycks2019augmix}. The trained network in this case has a clean test accuracy of $99\%$.

\end{enumerate}

\textbf{Memory Classifier: } The feature extractor in this case study predicts the color of the image. For pre-processing, we partition the image into segments via the method from \cite{Felzenszwalb2004}. 
Out of the $T$ largest  segments, we find the one with the highest intensity over the different color channels. 
In our experiments we set $T = 20$. Among the segments, the color channel with the highest intensity assigns the color to an image.  
Clustering the dataset with this feature forms three  memories. The node classifiers for each memory cluster use the Augmix data-augmentation technique to train a ResNet50 architecture. 
This again achieves a test accuracy of $100\%$.

\begin{figure}[htb]
\centering
\includegraphics[height = 4.4cm, width=13cm]{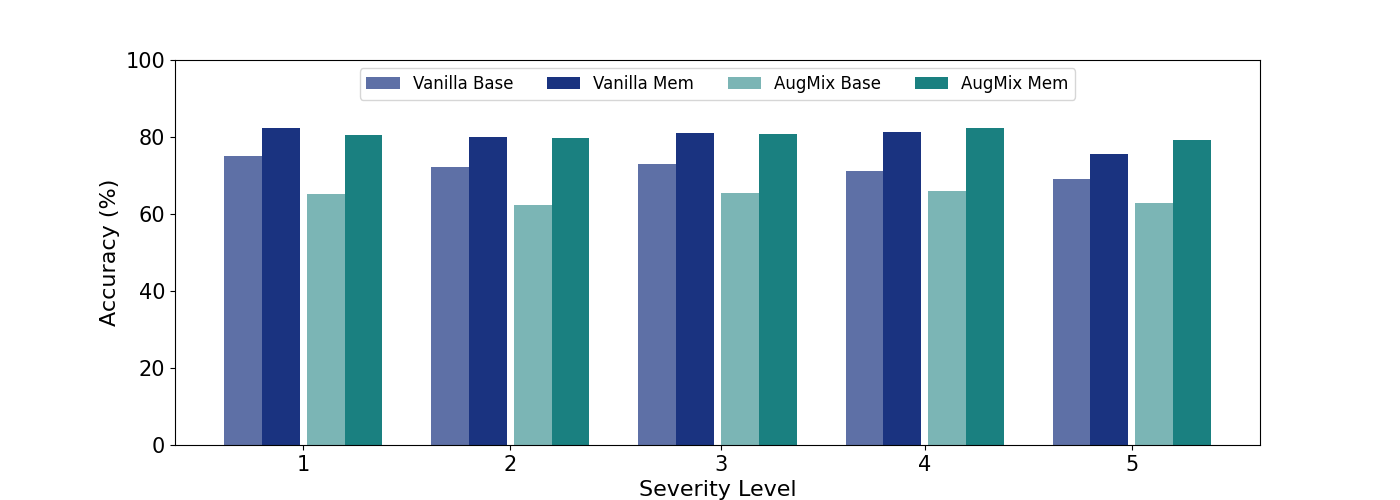}
\caption{\footnotesize Robust accuracy on corrupted color dataset across five severity levels }
\label{fig:color_data_aug_comp}
\end{figure}

\textbf{Results: } We subject this dataset to different perturbations and corruptions as discussed in \citep{hendrycks} and report the results in Table \ref{tab:color-severity-data-aug-results} and Figure \ref{fig:color_data_aug_comp}. 
We observe that memory classifiers are consistently more robust than ResNet50 with and without AugMix.

\begin{table*}[ht]
\tiny
\begin{center}
\begin{tabular}{ p{1.5cm} p{0.5cm} p{1cm} p{0.5cm} p{1cm} }
    \hline
    & \multicolumn{2}{ c }{Vanilla} &  \multicolumn{2}{c}{AugMix} \\
\hline 
& Baseline & Memory Classifier &  Baseline & Memory Classifier \\
    \hline 
    Clean Accuracy & 100 & 100 & 99 & \textbf{100}\\
    \hline
    \hline
    
Fog  & 35 & \textbf{50 }& 33 & \textbf{41} \\ 

Brightness  & 37 & \textbf{56} & 33 & \textbf{35} \\ 

Contrast  & 66 & \textbf{76} & 55 & \textbf{84} \\ 

Elastic  & 100 & 100 & 88 & \textbf{92} \\ 

Pixelate  & 100 & 100 & 91 & \textbf{94} \\ 

JPEG  & \textbf{97} & 96 & 84 & \textbf{94} \\ 
\rowcolor{blue!20}
Speckle Noise  & 99 & \textbf{100} & 79 & \textbf{95} \\ 
Gaussian Blur  & 100 & 100 & 89 & \textbf{94} \\ 
\rowcolor{blue!20}
Spatter  & 37 & \textbf{68} & 40 & \textbf{82} \\ 
Saturate  & 78 & \textbf{82} & 68 & \textbf{94} \\ 
\rowcolor{blue!20}
Gaussian Noise  & 40 & \textbf{65} & 46 & \textbf{95} \\ 
\rowcolor{blue!20}
Shot Noise  & 99 & \textbf{100} & 80 & \textbf{87} \\ 
\rowcolor{blue!20}
Impulse Noise  & 43 & \textbf{57} & 42 & \textbf{93} \\ 

Defocus Blur  & 100 & 100 & 87 & \textbf{94} \\ 

Zoom Blur  & 98 & 98 & 86 & \textbf{94} \\ 

Frost  & 35 & \textbf{53} & 33 & \textbf{41} \\ 

Motion Blur  & 100 & 100 & 89 & \textbf{94} \\ 

Snow  & 35 & \textbf{40} & 33 & \textbf{47} \\ 

\end{tabular}
\end{center}
\caption{\footnotesize\textit{Color data augmentation results.} We present the accuracies of classifiers for predicting the color across different corruptions and severities. The \textit{Baseline} and memory classifier uses a ResNet50 backbone with AugMix augmentation. 
We highlight the rows pertaining to different kinds of image noise.}
\label{tab:color-severity-data-aug-results}
\end{table*}

\subsection{Intializing Memories}

\textbf{Intializing Memories}: 
Certain  ``high-level" properties of images from the same class typically cluster well. For instance, the class of images containing a  red traffic light 
usually contain a set of bright red pixels 
Thus, we begin by efficiently identifying such broad-ranging categories.
While we know that having a small number of memories can help with generalization, the \emph{specific number} of memories is often not known a priori. 
We begin by placing memories in the input space densely enough, with the constraint that every training data point occurs within a similarity score threshold $\mathsf{b}_t$ of some memory. 

We demonstrate this in Algorithm \ref{alg:learn_init_memories}. A randomly picked data point is compared with all data points in the set \textit{RejectedSet}. This happens in a single linear pass. 
Data points marked as similar enough are included into the set of data points for the current  memory. 
Further analysis of these data points is not required for the current algorithm. 
This continues until there are any data points left. 
Algorithm \ref{alg:learn_init_memories} always terminates, as the size of \textit{RejectedSet} decreases in each iteration.
The output provides a warm start for any subsequent algorithm optimizing the choice of memories.

\label{sec:memory_initialization}
\begin{algorithm}[h]
\caption{Generate Initial Memories}
\label{alg:learn_init_memories}
\flushleft
\textbf{Input: } Data set $\mathcal{D}$:$ \{x_1, x_2, \dots, x_n \} $ \\
\textbf{Output: } Memories $ M:\{ m_1, m_2, m_3, \dots, m_q \}  $ \\
\textbf{Parameter}: Similarity Score Threshold $\mathsf{b}_t \in (0,1)$
\begin{algorithmic}[1]
\STATE $M = \phi$
\STATE RejectedSet = $\mathcal{D}$

\WHILE {RejectedSet  $\neq \phi$}
\STATE $m^*$ = selectRandomPoint(RejectedSet)
\FOR{$x_i \in $ RejectedSet}
\IF{$\score (m^*, x_i) > \mathsf{b}_t$}
\STATE RejectedSet = RejectedSet $ \setminus x_i$
\ENDIF
\ENDFOR
\STATE $M = M \cup m^*$
\ENDWHILE
\STATE \OUTPUT $M$
\end{algorithmic}
\end{algorithm}

\end{document}